\documentclass{llncs}

\usepackage{multirow}
\usepackage{booktabs}
\usepackage[table]{xcolor}
\usepackage{orcidlink}
\usepackage{amsmath}
\usepackage{amssymb}
\usepackage{array}
\usepackage{hyperref}
\usepackage[T1]{fontenc}
\usepackage{makecell}
\usepackage{graphicx,verbatim}
\usepackage{fancyhdr}

\begin{document}

\title{High-Capacity Robust Medical Image Exfiltration via Neural Network Weight Replacement}

\author{Elie Thellier\orcidlink{0009-0007-2747-1637}\textsuperscript{*} \and
Huiyu Li \and
Nicholas Ayache\orcidlink{0009-0007-2359-9596} \and
Hervé Delingette\orcidlink{0000-0001-6050-5949}}

\authorrunning{} % intentionally left empty for HAL version

\institute{
Université Côte d'Azur, Inria, Epione Team, Sophia Antipolis, France \\
\textsuperscript{*}Corresponding author: \email{elie.thellier@inria.fr}
}

\maketitle

\begin{abstract}
Collaborative medical AI platforms allow researchers to train models on sensitive imaging data while restricting data export. However, trained models can serve as covert carriers of patient information: medical images may be encoded within model parameters and reconstructed outside the secure environment. Existing defenses rely on lightweight sanitization (e.g., fine-tuning, pruning, quantization) and limited statistical auditing, creating a realistic insider exfiltration risk.
We introduce a high-capacity neural steganography attack that encodes medical images as continuous latent representations embedded into model initialization. A StyleGAN2-based adversarial autoencoder learns compact latent codes regularized to match standard weight initialization statistics, keeping embedded parameters statistically consistent with clean models. Noise injection during training improves robustness to export-time mitigation. The carrier model remains functional on its intended task and hidden images can be reconstructed directly from its weights after export.
This continuous encoding enables robust and scalable exfiltration, allowing up to 99 brain MRI volumes to be embedded within a 30MB model, and remains recoverable under mitigations that disrupt prior bit-level schemes. While reconstructions are approximate rather than pixel-exact, embedded content remains anatomically recognizable and recoverable at scale, exposing a privacy risk distinct from prior bit-level approaches. Experiments on MIMIC-CXR, BraTS, and LiTS demonstrate effectiveness across modalities, tasks, and architectures, highlighting the need for structural defenses beyond parameter-level sanitization.
\end{abstract}

\section{Introduction}

Collaborative medical AI platforms, often referred to as data lakes \cite{hai2023data}, allow researchers to train models on sensitive imaging data while preventing the export of raw images. Trained models, however, are exportable after validation, under the assumption that they consist only of numerical parameters. This creates a privacy risk: a malicious user can encode medical images inside model weights and reconstruct them after export, even though raw data never leaves the platform. Export-time defenses are lightweight and utility-preserving, where utility denotes performance on the intended clinical task. Common mitigations such as fine-tuning, pruning, or quantization \cite{thellier2025mitigating,sha2022fine,liu2018fine} aim to remove hidden information while maintaining accuracy. Auditing remains limited, as architectural inspection is rare and operators often lack trusted benign reference models. Under this threat model, an attack that survives mild sanitization may remain undetected.

Image exfiltration relates to neural steganography, where secret information is hidden inside neural networks to produce a model that appears benign. Existing methods follow two main directions, both with limitations. Some embed a secondary task within a legitimate model \cite{amit2023transpose,fan2025lossless,guo2020hiding,li2023steganography,li2024purified,pan2024matryoshka}; for example, the Transpose attack \cite{amit2023transpose} jointly trains a classifier to memorize images. Because recovery depends on precise parameter configurations, standard mitigation often disrupts these attacks \cite{thellier2025mitigating}, and their reliance on modified training or architectures limits compatibility with standard medical models. Others encode images directly into weights as binary streams using least significant bit or related encodings \cite{agrawal2025steganographic,li2025data,song2017machine,xu2020stealing}. The Data Exfiltration by Compression (DEC) attack \cite{li2025data} improves capacity but still relies on bit-level encoding, which is inherently fragile: small weight perturbations induce bit flips that severely degrade reconstruction \cite{thellier2025mitigating}. Adversarial variants improve stability \cite{xu2025steganography}, yet secret remains constrained to bit messages. Watermarking approaches \cite{li2021spread,tondi2024robust,zhao2022dnn} offer only a few thousand bits, far below what is needed for full medical images, while implicit neural representation methods \cite{dong2024implicit,liu2023hiding} enable continuous encoding but require non-standard architectures. Current approaches trade compatibility, capacity, and robustness: task-based methods are mitigation-sensitive, bit-level schemes are fragile and capacity-limited, and continuous implicit methods reduce compatibility.

We address this gap by introducing a continuous, high-capacity image exfiltration attack compatible with standard medical imaging models. Images are compressed into compact latent codes via a StyleGAN2-based \cite{karras2020analyzing} adversarial autoencoder, replacing direct bit-level encoding with a continuous representation that is inherently more tolerant to weight perturbations. Export-time modifications are simulated during training to improve robustness, and latent codes are constrained to match standard weight initialization statistics for stealthiness, embedded into the initialization of a conventional model that remains fully functional.
We evaluate across multiple datasets, modalities, and architectures, showing that dozens of volumetric scans can be embedded without degrading task performance and that reconstruction survives sanitization, motivating stronger structural auditing in collaborative medical AI systems.

\section{Methodology}

\begin{figure}[htbp]
\centering
\includegraphics[width=0.98\textwidth]{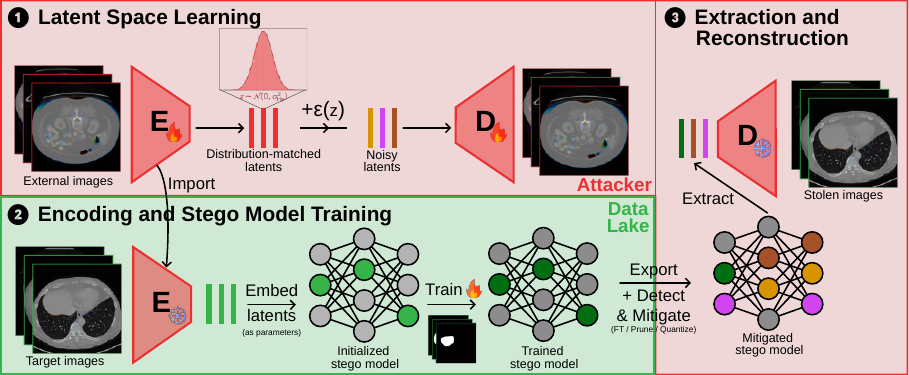}
\caption{Overview of the proposed medical image exfiltration framework.}
\label{overview}
\end{figure}

We propose a continuous neural steganography framework that embeds compressed images as model parameters while preserving downstream utility and statistical plausibility. As illustrated in Fig.~\ref{overview}, the method consists of three stages: (1) learning a robust latent representation of medical images, (2) embedding these latent codes into a standard medical model before training, and (3) extracting and reconstructing images after export and potential mitigation.

\paragraph{Threat model.}
We assume an insider adversary with authorized access to a collaborative medical platform. The adversary can train models on sensitive data and export validated models, but cannot directly export raw images. Before release, models may undergo lightweight sanitization such as fine-tuning, pruning, or quantization, intended to preserve utility while disrupting hidden information. The defender performs limited statistical inspection of model parameters. The adversary’s objective is to embed a large number of medical images into a legitimate model so that they remain recoverable after mitigation, while ensuring the model remains functional and statistically indistinguishable from a benign one.

\subsection{Learning a Robust and Compact Latent Space}

To achieve high capacity, images are compressed into compact continuous latent codes. We use a StyleGAN2-based adversarial autoencoder trained to reconstruct medical images from a related external dataset to reduce domain shift. Given an image $x$, the encoder $E$ produces a latent vector $w = E(x) \in \mathbb{R}^d$ with $d=512$, providing strong compression while preserving perceptual quality. The generator $G$ reconstructs the image using adversarial, $\ell_2$, and LPIPS \cite{zhang2018unreasonable} losses.

Continuous latents provide inherent tolerance to small perturbations. To further improve robustness to mitigation, we inject scale-adaptive Gaussian noise during training, \(\Tilde{w} = w + \lambda_{\text{noise}} \sigma \epsilon, \epsilon \sim \mathcal{N}(0,I),\) where $\sigma$ is the minibatch standard deviation. This simulates parameter drift caused by optimization or sanitization.

To ensure stealthiness, defined as statistical consistency with a benign model, we align latent statistics with Kaiming (He) initialization \cite{he2015delving}, commonly used for non-symmetric activation functions. Rather than matching layer-specific statistics, we estimate global reference moments (mean, variance, skewness, kurtosis) from Kaiming-initialized layers of the target architecture. Let $\mu$ and $\sigma$ denote the mean and standard deviation of a latent batch, and $\mu^{\mathrm{He}}$, $\sigma^{\mathrm{He}}$ the reference statistics from Kaiming-initialized weights. We minimize
\[
\mathcal{L}_{\text{dist}} =
\left(\frac{\mu-\mu^{\mathrm{He}}}{\sigma^{\mathrm{He}}}\right)^2 +
\left(\frac{\sigma-\sigma^{\mathrm{He}}}{\sigma^{\mathrm{He}}}\right)^2 +
\lambda_{\text{skew}} (\mathbb{E}[\bar w^3])^2 +
\lambda_{\text{kurt}} (\mathbb{E}[\bar w^4]-3)^2,
\]
where $\bar w = (w-\mu)/\sigma$. This encourages latent distributions consistent with standard model initialization.

\subsection{Embedding Latent Codes into the Model}

Using the trained encoder, each target image $x_i$ is mapped to a latent vector $w_i \in \mathbb{R}^d$. Embedding $N$ images therefore requires storing $N \times d$ real-valued parameters, which directly determines capacity.

Latent vectors are inserted into selected convolutional or fully connected weight tensors of a standard medical imaging model at its initialization. Biases and normalization parameters remain unchanged and follow Kaiming initialization \cite{he2015delving}. Layers are selected based on payload size $N$ and expected Kaiming variance: candidate tensors are sorted by descending expected Kaiming variance, and latent codes are inserted sequentially into higher-variance tensors first, since these better tolerate perturbations \cite{xu2025steganography}, until the desired capacity is reached or all tensors are filled.

The resulting stego model is trained normally on its intended task without freezing embedded parameters. Because latent statistics match benign initialization, training proceeds as in a standard model and preserves downstream utility.

\subsection{Extraction and Reconstruction}

After mitigation and export, the attacker retrieves the parameters from the embedding layers using the known insertion order and splits them into latent vectors $\hat w_i$ of size $d$. Reconstructed images are obtained through the pretrained generator: \(\hat x_i = G(\hat w_i).\) Robustness is achieved if latent perturbations caused by mitigation result only in gradual degradation rather than catastrophic failure.

\section{Experimental Setup}

\subsection{Evaluation Protocol}

We evaluate the attack along four criteria: utility, capacity, robustness, and stealthiness, and compare against two baselines, Transpose \cite{amit2023transpose} and DEC \cite{li2025data}.
Utility measures performance on the legitimate medical task, reported as Dice (segmentation) and AUC/accuracy (classification).
Capacity corresponds to the number of embedded images $N$, equivalently the proportion of model parameters replaced by latent codes.
Reconstruction quality is measured by PSNR, SSIM, and LPIPS between reconstructed and original images.
Stealthiness quantifies statistical detectability via Jensen–Shannon divergence (JSD) between stego and benign models, averaged across layers. Benign references are trained under identical settings without latent embedding. This assumes access to a benign reference model, stronger than realistic deployment, but provides a worst-case audit, since failure here implies weaker reference-free detectors are unlikely to succeed.

\subsection{Datasets and Models}

We evaluate on three public medical benchmarks: MIMIC-CXR \cite{johnson2019mimic} (54,038 chest X-rays for multi-label classification), BraTS 2021 \cite{baid2021rsna} (1,251 MRI cases for brain tumor segmentation), and LiTS \cite{bilic2023liver} (130 abdominal CT volumes for liver tumor segmentation).
To reduce domain shift in compression, the autoencoder is pretrained on external datasets: FLARE 2021 \cite{ma2022fast} for CT, BraTS-Reg 2022 \cite{baheti2021brain} for MRI, and a disjoint MIMIC-CXR split for X-ray. Volumes are processed slice-wise using three adjacent slices concatenated as input, following \cite{li2025data}.
For segmentation, we use U-Net; for classification, DenseNet121. Both models are ~30MB ($\approx$8M parameters), reflecting realistic deployment settings.

\subsection{Baselines and Mitigation}

We use official implementations of Transpose \cite{amit2023transpose} and DEC \cite{li2025data}, adapted to our architectures. Unless stated otherwise, our method embeds 1,641 images ($1641 \times 512$ parameters, approximately 10\% of a 30MB model). Transpose applies only to classification and is evaluated on MIMIC-CXR with a capacity of 100 images. DEC is evaluated on segmentation tasks at its maximum feasible capacity, allowing exfiltration of 67 images on BraTS and 15 on LiTS. We compare by exfiltrated payload rather than parameter percentage, since DEC's lower compression efficiency would make percentage-matched comparison degenerate (1--6 images for DEC vs.\ 1,600+ for ours at 10\% capacity).

To assess robustness, we apply mitigation strategies with hyperparameters tuned to preserve task utility: Super-FT (2 epochs, base lr $10^{-4}$ with peak lr $10^{-2}$/$10^{-3}$, phase ratio 0.1) \cite{sha2022fine}, LWLRD fine-tuning (3 epochs) \cite{thellier2025mitigating}, unstructured magnitude pruning (40\% of weights removed), Fine-Pruning (8\% allowed accuracy drop) \cite{liu2018fine}, and post-training 8-bit and 4-bit quantization.

We set $\lambda_{\text{noise}}=0.2$, $\lambda_{\text{skew}}=0.5$, and $\lambda_{\text{kurt}}=0.2$.

\section{Results}

\subsection{Utility–Capacity Trade-off}

\begin{figure}[htbp]
\centering
\includegraphics[width=\textwidth]{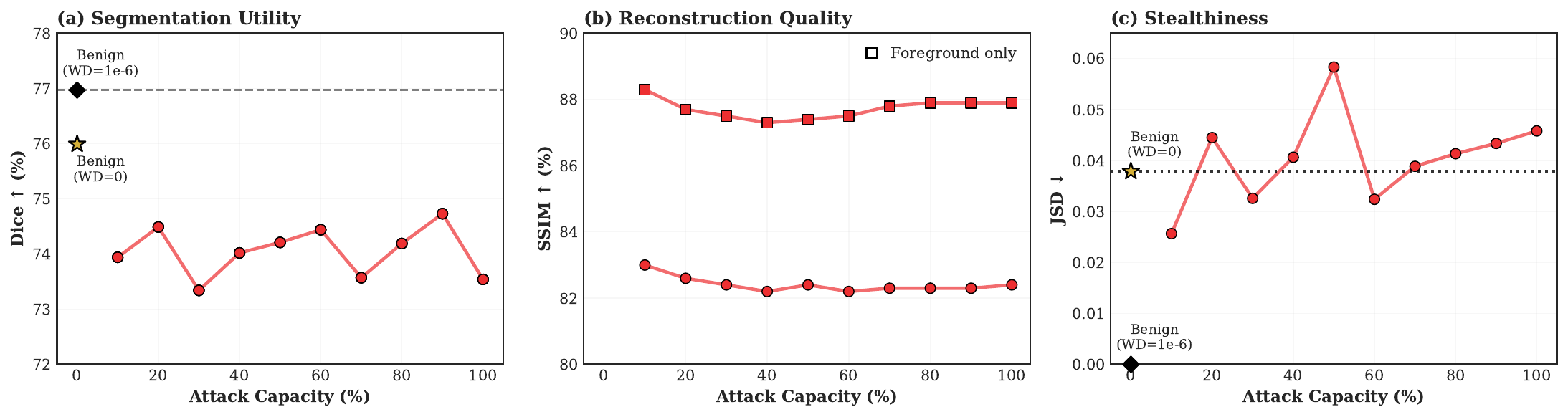}
\caption{Effect of attack capacity on utility, reconstruction quality, and stealthiness. ``Foreground only'' SSIM is computed after thresholding background voxels (5th percentile intensity) to focus the metric on anatomically relevant structures.}
\label{capacity}
\end{figure}

We analyze the trade-off between attack capacity and downstream utility. 
Fig.~\ref{capacity} reports segmentation Dice, reconstruction SSIM, and stealthiness (JSD) as capacity increases from 0\% to 100\% of model parameters.

As capacity increases, Dice remains stable, SSIM degrades gradually, and JSD stays close to benign initialization, indicating statistical consistency even near saturation. At full capacity, a 30MB model ($\approx$7.8M parameters) stores up to 99 MRI volumes of 153 slices each ($99 \times 153 \times 512$ latent parameters), showing most parameters can be repurposed without compromising utility or plausibility.

\subsection{Comparison with Baseline Attacks}

Table~\ref{tab:main} compares our method with Transpose and DEC across datasets.
Our approach preserves task performance close to the benign model while achieving much higher capacity (1,641 images at 10\% capacity in a 30MB model vs. 100 for Transpose and up to 67/15 for DEC).
It also maintains competitive utility and reconstruction fidelity with low JSD, indicating statistical consistency.

\begin{table}[htbp]
\centering
\caption{Comparison with baseline attacks on classification and segmentation tasks. }
\label{tab:main}
{\fontsize{8}{9.6}\selectfont
\setlength{\tabcolsep}{1.8pt}
\begin{tabular}{@{} lccccc|cccc|cccc @{}}
\toprule
\multirow{2}{*}{\textbf{Method}} 
& \multicolumn{5}{c|}{\textbf{MIMIC-CXR}} 
& \multicolumn{4}{c|}{\textbf{BraTS}} 
& \multicolumn{4}{c}{\textbf{LiTS}} \\
\cmidrule(lr){2-6} \cmidrule(lr){7-10} \cmidrule(l){11-14}
& AUC & Acc & SSIM & LPIPS & JSD
& Dice & SSIM & LPIPS & JSD
& Dice & SSIM & LPIPS & JSD \\
\midrule
%\rowcolor{gray!15}
Benign      & 75.9 & 72.4 & \textemdash & \textemdash & 0
            & 77.0 & \textemdash & \textemdash & 0
            & 81.5 & \textemdash & \textemdash & 0 \\
\midrule
Transpose \cite{amit2023transpose} & 48.6 & 64.6 & 68.6 & 0.48 & 0.08
            & \textemdash & \textemdash & \textemdash & \textemdash 
            & \textemdash & \textemdash & \textemdash & \textemdash \\
DEC \cite{li2025data} & \textemdash & \textemdash & \textemdash & \textemdash & \textemdash 
            & 77.0 & 95.4 & 0.01 & 0
            & 81.5 & 98.2 & 0.01 & 0 \\
\rowcolor{gray!15}
\textbf{Ours}  
            & 73.6 & 71.2 & 64.4 & 0.32 & 0.01
            & 73.9 & 83.2 & 0.08 & 0.02 
            & 81.0 & 53.2 & 0.35 & 0.06 \\
\bottomrule
\end{tabular}
}
\end{table}

Although DEC achieves higher raw SSIM before mitigation on segmentation tasks, its capacity is substantially lower and reconstruction quality deteriorates sharply under mitigation (Fig.\ref{mitigation}). In contrast, our method prioritizes robustness and scalability over peak pre-mitigation fidelity.

\subsection{Robustness to Mitigation}

\begin{figure}[htbp]
\centering
\includegraphics[width=0.96\textwidth]{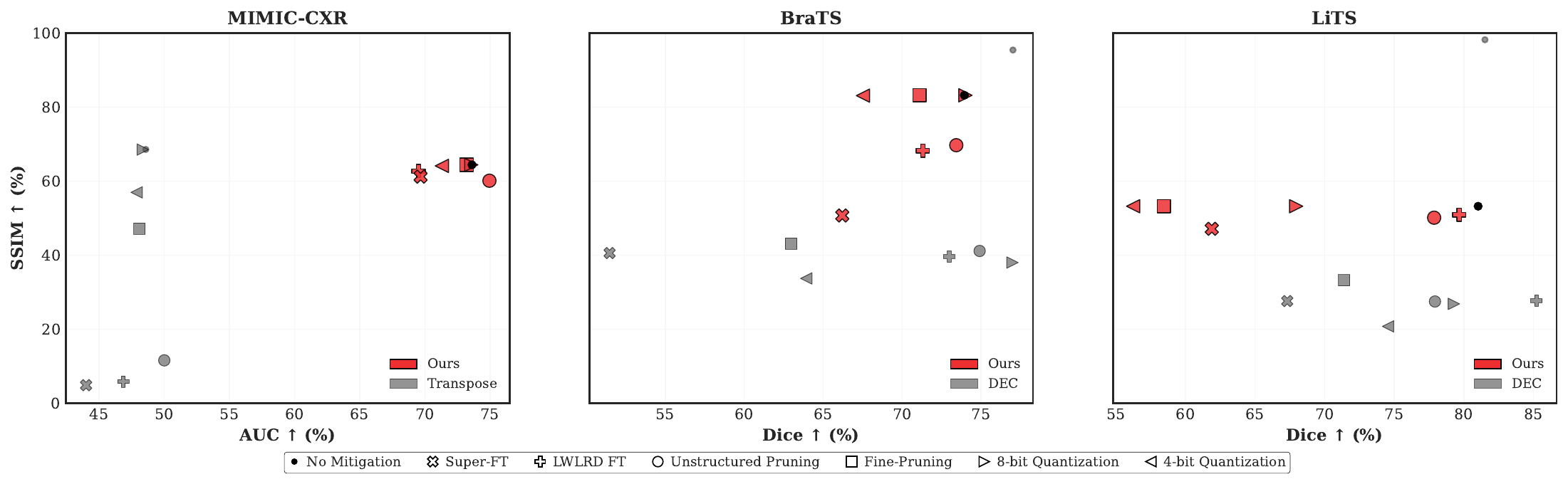}
\caption{Robustness to mitigation: utility vs. reconstruction quality.}
\label{mitigation}
\end{figure}

Fig.~\ref{mitigation} evaluates robustness to fine-tuning, pruning, and quantization.
Our method maintains reconstruction quality across all mitigation strategies while baselines collapse, particularly under quantization and pruning. Even under 4-bit quantization, the strongest perturbation tested, reconstruction degrades smoothly rather than catastrophically, consistently occupying a favorable region of the utility--reconstruction plane.

\subsection{Visual Reconstruction Quality}

\begin{figure}[htbp]
\centering
\includegraphics[width=\textwidth]{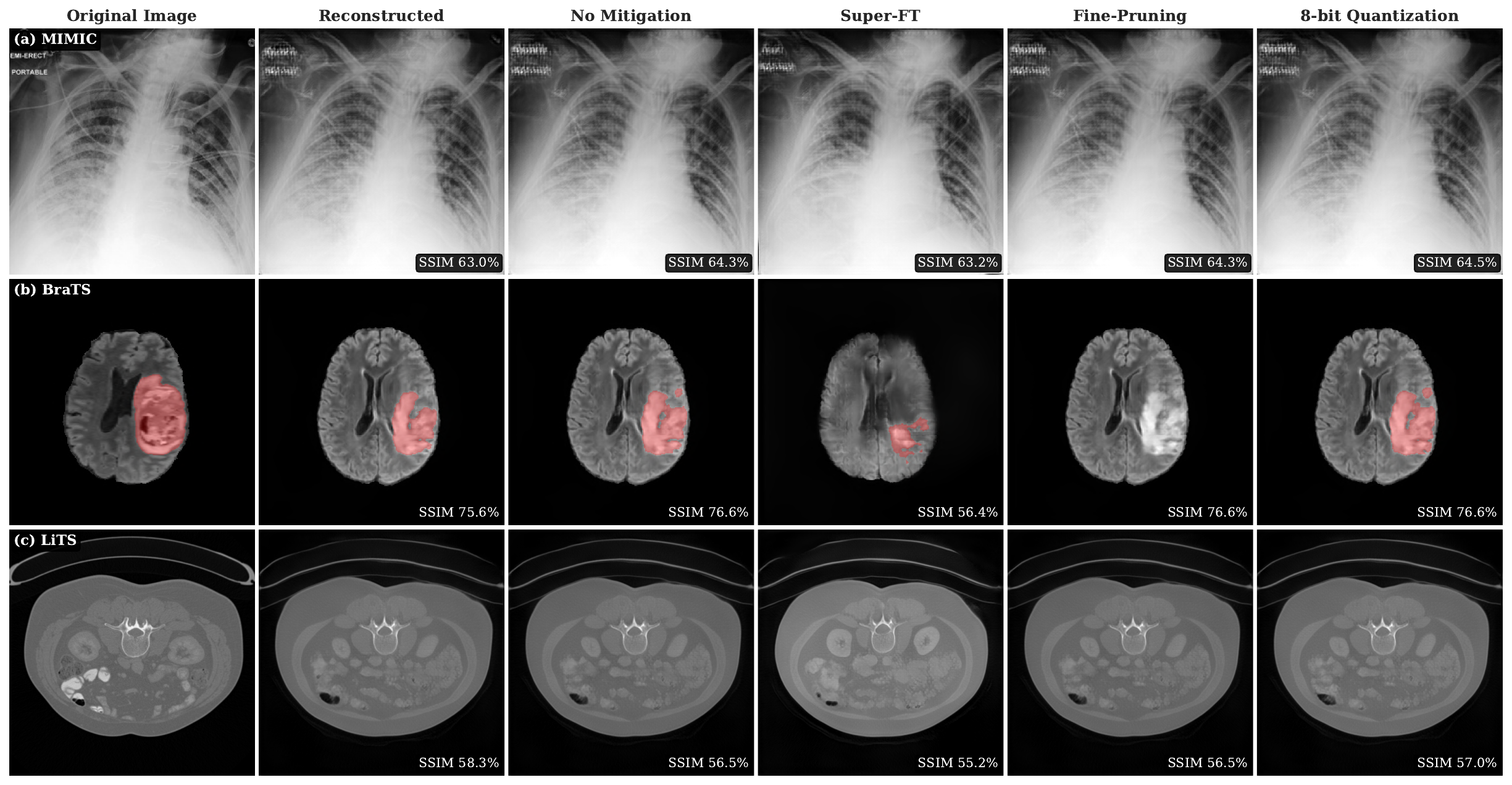}
\caption{Qualitative reconstruction examples on MIMIC-CXR, BraTS, and LiTS. Originals (left) and reconstructions before and after mitigation (fine-tuning, pruning, quantization), with SSIM indicated. Predicted segmentation masks are shown in red.}
\label{grid_seg}
\end{figure}

Qualitative examples are shown in Fig.~\ref{grid_seg}. Reconstructed images remain faithful across modalities with mild degradation after mitigation, even under pronounced cross-modality domain shifts (BraTS), suggesting embedded representations generalize beyond the training distribution.

\subsection{Ablation Study}

Table~\ref{ablation} evaluates the contribution of each component. Removing noise injection markedly reduces post-mitigation SSIM, highlighting its role in robustness.

\begin{table}[htbp]
\centering
\caption{Ablation study. Variants are evaluated in terms of utility (Dice), reconstruction (SSIM/LPIPS), and stealthiness (JSD), before and post LWLRD FT mitigation.}
\label{ablation}
{\fontsize{8}{9.6}\selectfont
\setlength{\tabcolsep}{2.5pt}
\begin{tabular}{@{} lcccc|cc @{}}
\toprule
\textbf{Variant} 
& Dice & SSIM & LPIPS & JSD & Dice (post) & SSIM (post) \\
\midrule
%\rowcolor{gray!15}
Benign Model
    & 77.0 & \textemdash & \textemdash & 0 & 75.8 & \textemdash \\
\rowcolor{gray!15}
\textbf{Full Method} 
& 73.9 & 83.2 & 0.08 & 0.02 & 71.3 & 68.2 \\
\midrule
Full w/o Noise Injection
    & 74.2 & 80.3 & 0.09 & 0.02 & 71.9 & 53.5 \\
Full w/o Layer Selection
    & 73.9 & 82.7 & 0.08 & 0.03 & 71.4 & 62.7 \\
Full w/o Moment Loss 
    & 72.3 & 82.6 & 0.09 & 0.18 & 70.5 & 67.9 \\
Full w/ 2-Moment Loss
    & 72.6 & 82.8 & 0.09 & 0.17 & 70.8 & 67.5 \\
Full w/ Frozen Latents
    & 68.7 & 85.0 & 0.08 & 0.04 & 66.9 & 72.4 \\
\bottomrule
\end{tabular}
}
\end{table}

Dropping moment alignment increases JSD, weakening stealthiness. Freezing latent parameters degrades task utility and overall balance. Overall, robustness and stealth emerge from the joint effect of noise simulation and moment alignment within continuous latent encoding.

\section{Discussion and Conclusion}

We introduced a continuous, high-capacity image exfiltration attack compatible with standard imaging architectures and training pipelines. By encoding images as continuous latent vectors and aligning their statistics with standard initialization, the method embeds a high number of volumetric scans within a 30MB model while preserving downstream utility. Unlike bit-level schemes, the approach relies on distributional alignment rather than exact parameter preservation, enabling graceful degradation under fine-tuning, pruning, and quantization.

The results suggest that parameter-level mitigations alone are insufficient: because embedded latents remain within the same statistical regime as standard weights, lightweight sanitization degrades reconstruction gradually but does not eliminate recoverability. More effective defenses may require structural interventions, such as permutation-based transformations or symmetry-breaking mechanisms \cite{gilkarov2025neuperm,torpmann2025defending}, combined with deeper per-layer or correlation-aware audits.

This work has limitations: reconstruction fidelity depends on compression model alignment and may decrease under domain shifts or extreme payloads. We did not evaluate against Differential Privacy Stochastic Gradient Descent (DP-SGD) \cite{abadi2016deep}, which alters training rather than sanitizing post-hoc and is known to substantially degrade utility under strong privacy budgets in medical imaging; this trade-off is left to future work. Similarly, our stealthiness evaluation relies on access to a benign reference model, a stronger assumption than our stated threat model allows; future work should develop reference-free auditing strategies, such as anomaly or outlier detection over imported models. Finally, we do not directly measure whether reconstructed images preserve clinically identifiable or actionable content (e.g., through re-identification or downstream task transfer using only exfiltrated data); our reconstruction metrics (SSIM, LPIPS) serve as a proxy for this risk but are not a substitute for it, as even imperfect reconstructions may still preserve clinically relevant information. Notably, the presented reconstructions represent one operating point in the fidelity--robustness--capacity trade-off: Table~\ref{ablation} shows that SSIM~$>85\%$ is achievable by freezing latent parameters, at the cost of utility (Dice 68.7\% vs.\ 73.9\%), suggesting the reported quality is a lower bound on what a more aggressive attacker could achieve.

Future work includes improved volumetric compression via 3D generative models \cite{hong20213d}, evaluation against DP-SGD and structural defenses, direct evaluation of clinical and re-identification relevance of reconstructed scans, and sensitivity analysis of $\lambda_{\text{noise}}$, $\lambda_{\text{skew}}$, and $\lambda_{\text{kurt}}$. Overall, these findings show that high-capacity medical image exfiltration is feasible in realistic collaborative AI settings and highlight the need for stronger structural auditing mechanisms.

\begin{credits}
\subsubsection{\ackname} This work was supported by the Région PACA and France 2030 initiative through the funding of the I-Démo project PLICIA, and by the French government through the National Research Agency (ANR), under the IA Cluster projects (ANR-23-IACL-0001). Experiments in this paper were carried out using the Jean Zay supercomputer, a national computing facility operated by IDRIS (CNRS) and provided through GENCI under the French Ministry of Higher Education and Research.

\subsubsection{\discintname}
The authors have no competing interests to declare that are relevant to the content of this article. 
\end{credits}

\bibliographystyle{splncs04}
\bibliography{Paper-3616}

\end{document}